\pdfoutput=1

\documentclass[11pt]{article}

\usepackage[final]{acl}

\usepackage{times}
\usepackage{latexsym}

\usepackage[T1]{fontenc}

\usepackage[utf8]{inputenc}

\usepackage{microtype}

\usepackage{inconsolata}

\usepackage{graphicx}
\usepackage{booktabs}    
\usepackage{makecell}    
\usepackage{array}       
\usepackage{adjustbox}   
\usepackage{multirow}
\usepackage{tcolorbox}
\usepackage{graphicx}   

\title{Spurious Correlations and Beyond: Understanding and Mitigating Shortcut Learning in SDOH Extraction with Large Language Models}

\author{
  \textbf{Fardin Ahsan Sakib}$^{1}$,
  \textbf{Ziwei Zhu}$^{1}$,
  \textbf{Karen Trister Grace}$^{2}$,
  \textbf{Meliha Yetişgen}$^{3}$,
  \textbf{Özlem Uzuner}$^{4}$
\\
  $^{1}$Department of Computer Science, $^{2}$School of Nursing, $^{4}$Department of Information Sciences and Technology \\
  George Mason University \\
  $^{3}$Department of Biomedical Informatics \& Medical Education, University of Washington \\
  \texttt{\{fsakib,zzhu20,kgrace,ouzuner\}@gmu.edu, melihay@uw.edu}
}

\begin{document}
\maketitle
\begin{abstract}
Social determinants of health (SDOH) extraction from clinical text is critical for downstream healthcare analytics. Although large language models (LLMs) have shown promise, they may rely on superficial cues leading to spurious predictions. Using the MIMIC portion of the SHAC (Social History Annotation Corpus) dataset and focusing on drug status extraction as a case study, we demonstrate that mentions of alcohol or smoking can falsely induce models to predict current/past drug use where none is present, while also uncovering concerning gender disparities in model performance. We further evaluate mitigation strategies—such as prompt engineering and chain-of-thought reasoning—to reduce these false positives, providing insights into enhancing LLM reliability in health domains.
\end{abstract}

\section{Introduction}
SDOH---including substance use, employment, and living conditions---strongly influence patient outcomes and clinical decision-making \cite{daniel2018addressing, himmelstein2018determined, armour2005annual}. Extracting SDOH information from unstructured clinical text is increasingly important for enabling downstream healthcare applications and analysis \cite{jensen2012mining, demner2009can}. Although LLMs have shown promise in clinical natural language processing (NLP) tasks \cite{hu2024improving, liu2023deid, singhal2023large}, they often rely on superficial cues \cite{tang2023large, zhao2017men}, potentially leading to incorrect predictions undermining trust and utility in clinical settings. 

Recent work has highlighted how LLMs can exhibit "shortcut learning" behaviors \cite{tu2020empirical,ribeiro2020beyond,zhao2018gender}, where they exploit spurious patterns in training data rather than learning causal, generalizable features. This phenomenon spans various NLP tasks, from natural language inference \cite{mccoy2019right} to question-answering \cite{jia2017adversarial}, and in clinical domains can lead to incorrect assumptions about patient conditions \cite{brown2023detecting,jabbour2020deep}, threatening the utility of automated systems.

We investigate how LLMs produce spurious correlations in SDOH extraction through using drug status time classification (current, past, or none/unknown) as a case study. Using the MIMIC \cite{johnson2016mimic} portion of the SHAC \cite{lybarger2021annotating} dataset, we examine zero-shot and in-context learning scenarios across multiple LLMs (Llama \cite{llama3_1_model_card}, Qwen \cite{yang2024qwen2technicalreport}, Llama3-Med42-70B \cite{christophe2024med42v2suiteclinicalllms}).We explore multiple mitigation strategies to address these spurious correlations: examining the causal role of triggers through controlled removal experiments, implementing targeted prompt engineering approaches like chain-of-thought (CoT) reasoning \cite{wei2022chain}, incorporating warning-based prompts, and augmenting with additional examples. While these interventions show promise—significant false positive rates persist, highlighting the deep-rooted nature of these biases and the need for more sophisticated solutions.

\noindent\textbf{Contributions:}
\begin{enumerate}
\item We present the first comprehensive analysis of spurious correlations in SDOH extraction across multiple LLM architectures, including domain-specialized models. Through extensive experiments in zero-shot and ICL settings, we demonstrate how models rely on superficial cues and verify their causal influence through controlled ablation studies.

\item We uncover systematic gender disparities in model performance, demonstrating another form of spurious correlation where models inappropriately leverage patient gender for drug status time classification predictions.


\item We evaluate multiple prompt-based mitigation strategies (CoT, warnings, more examples) and analyze their limitations, demonstrating that while they reduce incorrect drug status time predictions, more robust solutions are needed for reliable clinical NLP deployments.
\end{enumerate}

\section{Related Work}
Previous work on extracting SDOH from clinical text spans a progression from rule-based methods to fine-tuned neural models, leveraging annotated corpora for tasks like substance use and employment status extraction \cite{hatef2019assessing, patra2021extracting, yu2022study, han2022classifying, uzuner2008identifying,stemerman2021identification, lybarger2023leveraging}. More recent efforts have explored prompt-based approaches with LLMs, including GPT-4, to reduce reliance on extensive annotations \cite{ramachandran2023promptbasedextractionsocialdeterminants}. While these approaches achieve competitive performance, studies across NLP tasks have shown that both fine-tuned and prompting-based methods often exploit spurious correlations or superficial cues \cite{ribeiro2020beyond, geirhos2020shortcut, tu2020empirical}. Prior investigations have focused largely on spurious correlations in standard NLP tasks and supervised scenarios \cite{mccoy2019right,zhao2018gender}. In contrast, our work examines how these issues manifest in zero-shot and in-context SDOH extraction settings, and we propose prompt-level strategies to mitigate these correlations.

\section{Methodology}



\subsection{Dataset and Task}


We use the MIMIC-III portion of the SHAC dataset~\cite{lybarger2021annotating}, which comprises 4405 deidentified social history note sections derived from MIMIC-III~\cite{johnson2016mimic} and the University of Washington clinical notes. SHAC is annotated using the BRAT tool~\cite{stenetorp2012brat}, capturing a variety of SDOH event types (e.g., Alcohol, Drug, Tobacco) as triggers along with associated arguments, including temporal status. To enable demographic analysis, we augmented the SHAC data by linking it with patient demographic information available in the original MIMIC-III dataset. 

In this work, we examine spurious correlations in SDOH extraction through temporal drug status classification (current, past, or none/unknown). We adopt a two-step pipeline \cite{ma2022dice, ma2023large}:

\textbf{(1) Trigger Identification:} Given a social history note, the model identifies spans corresponding to the target event type (e.g.,drug use).


\textbf{(2) Argument Resolution:} For each trigger, models applies a multiple-choice QA prompt to determine the temporal status (current/past/none).  


The dataset contains diverse patterns of substance documentation, see Appendix~\ref{sec:appendix-dataset} for detailed examples of the task and annotation schema. 


\subsection{Experimental Setup}
\paragraph{Model Configurations} We evaluate multiple model configurations:
\begin{itemize}
    \item \textbf{Zero-Shot:} Models receive only task instructions and input text, with no examples
    \item \textbf{In-Context Learning (ICL):} Models are provided with three example demonstrations before making predictions on a new instance. Examples are selected to maintain balanced representation across substance use patterns (none/single/multiple) and drug use outcomes (positive/negative).
    
    \item \textbf{Fine-Tuning (SFT):} We also fine-tune a Llama-3.1-8B model on the MIMIC portion of the SHAC dataset to assess whether domain adaptation reduces spurious correlations.
    
\end{itemize}

We consider Llama-3.1-70B (zero-shot, ICL), Llama-3.1-8B (fine-tuned on MIMIC), Qwen-72B (ICL), and Llama3-Med42-70B (ICL). These models span various parameter sizes and domain specializations. The fine-tuned Llama-8B model provides insights into whether in-domain adaptation mitigates the observed shortcut learning.

\paragraph{Prompting Strategies} We additionally use two other prompting strategies (see Appendix~\ref{sec:prompting_strategies} for complete templates):

\textbf{Chain-of-Thought (CoT):} This prompt explicitly guides reasoning through five steps: (1) read the social history note carefully, (2) identify relevant information, (3) consider examples provided, (4) explain your reasoning process, (5) provide the answer. This encourages explicit reasoning to reduce shortcuts.

\textbf{Warning-Based:} This incorporates explicit guidelines to counter spurious correlations: (1) evaluate each factor independently - never assume one behavior implies another, (2) extract only explicitly stated information - avoid making assumptions based on demographic or other factors, (3) use [none] when information isn't mentioned.

\paragraph{Evaluation Framework}
Our primary evaluation metric is the false positive rate (FPR), defined as:
$FPR = FP / (FP + TN)$
where FP represents false positives (predicted current/past use when ground truth was none/unknown) and TN represents true negatives (correctly predicted none/unknown). We prioritize FPR given the clinical risks of incorrect positive drug use predictions—including patient stigmatization, biased provider perceptions, and diminished trust in automated systems~\cite{van2013stigma, dahl2022investigating}. A higher FPR indicates more frequent erroneous predictions that could directly impact patient care. We specifically examine FPR disparities between substance-positive and substance-negative contexts to reveal whether models rely on superficial cues rather than actual evidence. See Appendix~\ref{sec:justification} for extended discussion.

To analyze potential spurious correlations, we categorize notes based on their ground truth substance use status:
\begin{itemize}
    \item \textbf{Substance-positive}: Notes documenting current/past use of the respective substance (alcohol or smoking)
    \item \textbf{Substance-negative}: Notes where the ground truth indicates no use or unknown status
\end{itemize}

\paragraph{Experimental Settings}
\begin{itemize}
    \item \textbf{Original:} Evaluate models on the original notes.
    \item \textbf{Without Alcohol/Smoking Triggers:} Remove mentions of alcohol/smoking to test their causal role in inducing false positives.
\end{itemize}

\section{Results}


\subsection{RQ1: Do Large Language Models Exhibit Spurious Correlations in SDOH Extraction?}


\begin{table*}[t]
    \centering
    \setlength{\tabcolsep}{1.5pt}
    \small
    \caption{False Positive Rates (\%) Across Different Models and Approaches. *Smoking+Alcohol* refers to cases where both *Smoking-positive* and *Alcohol-positive* are true.}

    \label{table1}
    \begin{tabular}{l|ccccc|cc|c|c}
        \toprule
        \multirow{2}{*}{Cases} & \multicolumn{5}{c|}{Llama-70B} & \multicolumn{2}{c|}{Llama-8B} & Llama3-Med42-70B & Qwen-72B \\
        \cmidrule{2-10}
        & Zero-shot & ICL & CoT & Warning & Increased-Examples & Vanilla & Fine-tuned & ICL & ICL \\
        \midrule
        Alcohol-positive & 66.21 & 48.28 & 33.79 & 40.69 & 45.52 & 73.10 & 32.41 & 66.90 & 62.76 \\
        Smoking-positive & 61.11 & 36.42 & 25.93 & 29.63 & 30.25 & 74.07 & 36.42 & 57.41 & 53.09 \\
        Alcohol-negative & 28.83 & 11.71 & 6.76 & 5.41 & 10.81 & 37.39 & 12.16 & 16.22 & 46.85 \\
        Smoking-negative & 29.76 & 18.05 & 10.73 & 11.22 & 20.00 & 33.66 & 7.32 & 19.51 & 53.17 \\
        Smoking+Alcohol & 73.26 & 51.16 & 34.88 & 45.35 & 39.53 & 81.40 & 40.70 & 76.74 & 56.98 \\
        \bottomrule
    \end{tabular}
\end{table*}



As shown in Table~\ref{table1}, our analysis in a zero-shot setting with Llama-70B reveals high false positive rates for drug status time classification in alcohol-positive (66.21\%) and smoking-positive (61.11\%) notes. In contrast, alcohol-negative and smoking-negative notes show substantially lower false positive rates (28.83\% and 29.76\%, respectively). This stark contrast suggests that the mere presence of alcohol or smoking triggers biases the model towards inferring nonexistent drug use. These biases likely stem from the pre-training phase, potentially reinforcing societal assumptions about correlations between different types of substance use.


\subsection{RQ2: Do In-Context Learning and Fine-Tuning Reduce These Spurious Correlations?}

Providing three in-context examples reduces false positives significantly. For Llama-70B, ICL lowers alcohol-positive mismatches from 66.21\% to 48.28\%, though a gap remains relative to alcohol-negative notes (11.71\%). Similarly, smoking-positive mismatches decrease from 61.11\% to 36.42\% versus 18.05\% for smoking-negative. The effectiveness of ICL suggests that explicit examples help the model focus on relevant features, though the persistence of some bias indicates deep-rooted associations from pre-training. Fine-tuning Llama-8B on the MIMIC subset (SFT) yields further improvements: alcohol-positive mismatches drop to 32.41\% and smoking-positive to 36.42\%, with corresponding negatives at ~12\% and ~7\% respectively, indicating that domain adaptation helps override some pre-trained biases.

\subsection{RQ3: Are These Superficial Mentions Causally Driving the Model's Predictions?}

To confirm the causal role of alcohol and smoking mentions, we remove these triggers from the notes. Across models, this consistently lowers false positives. For instance, Llama-70B zero-shot sees alcohol-positive mismatches fall from 66.21\% to 55.17\% after removing alcohol triggers. Similarly, Llama-8B-SFT reduces alcohol-positive errors from 32.41\% to 26.9\%. Similar trends are observed across other architectures including domain-specific models (see appendix~\ref{sec:trigger-removal}), confirming that alcohol and smoking cues spuriously bias the models' drug-use predictions.

\subsection{RQ4: Are there systematic demographic variations in these spurious correlations?}

Beyond substance-related triggers, our analysis (Table~\ref{table2}) uncovers another concerning form of spurious correlation: systematic performance differences based on patient gender. Just as models incorrectly rely on mere mentions of alcohol or smoking to infer substance use, they appear to leverage patient gender as an inappropriate predictive signal. For the base Llama-70B model in zero-shot settings, false positive rates show stark gender disparities - male patients consistently face higher misclassification rates compared to female patients (71.15\% vs 53.66\% for alcohol-positive cases, and 66.67\% vs 50.88\% for smoking-positive cases). This pattern persists with in-context learning, with the gender gap remaining substantial (alcohol-positive: 52.88\% male vs 36.59\% female). Fine-tuned models showed similar disparities, with Llama-8B-SFT maintaining a performance gap of approximately 15 percentage points between genders for alcohol-positive cases.

Notably, these gender-based differences exhibit complex interactions with substance-related triggers. Cases involving positive substances mentions show the most pronounced disparities, with male patients seeing up to 20 percentage point higher false positive rates. This suggests that the model's shortcut learning compounds across different dimensions - gender biases amplify substance-related biases and vice versa. The persistence of these interacting biases across model architectures, sizes, and prompting strategies suggests they arise from deeply embedded patterns in both pre-training data and medical documentation practices.

\section{Mitigation Strategies and Results}
We explore several mitigation techniques to address the spurious correlations identified in our analysis:
\paragraph{Chain-of-Thought (CoT)}
As shown in Table~\ref{table1}, instructing the model to reason step-by-step before producing an answer leads to substantial reductions across all architectures. For Llama-70B, CoT reduces alcohol-positive mismatches from 66.21\% (zero-shot) to 33.79\%, with smoking-positive cases decreasing from 61.11\% to 25.93\%. Similar improvements are observed in other models (see appendix~\ref{sec:mitigation-experiments}), with Qwen-72B showing particularly strong response to CoT. This suggests CoT helps models avoid superficial cues and focus on explicit information.
\paragraph{Warning-Based Instructions}
We prepend explicit instructions cautioning the model not to assume drug use without evidence and to treat each factor independently. With Llama-70B, these warnings lower alcohol-positive mismatches from 66.21\% to approximately 40.69\%, and also benefit smoking-positive scenarios. While not as strong as CoT, these warnings yield meaningful improvements across different architectures.
\paragraph{Increased Number of Examples}
Providing more than three examples---up to eight---further stabilizes predictions. For Llama-70B, increasing the number of examples reduces false positive rates considerably, with alcohol-positive mismatches falling to 45.52\% (compared to 66.21\% zero-shot). Similar trends are observed in other models, though the magnitude of improvement varies (see appendix~\ref{sec:mitigation-experiments}). While not as dramatic as CoT, additional examples help guide models away from faulty heuristics.



\begin{table*}[t]
    \centering
    \setlength{\tabcolsep}{4pt}
    \small
    \caption{Gender-Based Analysis of False Positive Rates (\%) Across Models}
    \label{table2}
    \begin{tabular}{l|cc|cc|cc|cc}
        \toprule
        & \multicolumn{2}{c|}{Llama-70B Zero-shot} & \multicolumn{2}{c|}{Llama-70B ICL} & \multicolumn{2}{c|}{Llama-8B SFT} & \multicolumn{2}{c}{Qwen-72B} \\
        \cmidrule{2-9}
        Cases & Female & Male & Female & Male & Female & Male & Female & Male \\
        \midrule
        Alcohol-positive & 53.66 & 71.15 & 36.59 & 52.88 & 21.95 & 36.54 & 68.29 & 60.58 \\
        Smoking-positive & 50.88 & 66.67 & 28.07 & 40.95 & 24.56 & 42.86 & 49.12 & 55.24 \\
        Alcohol-negative & 29.13 & 28.42 & 9.45 & 14.74 & 9.45 & 15.79 & 47.24 & 46.32 \\
        Smoking-negative & 27.03 & 32.98 & 9.91 & 27.66 & 6.31 & 8.51 & 54.05 & 52.13 \\
        Smoking+Alcohol & 81.82 & 84.62 & 54.55 & 58.97 & 27.27 & 53.85 & 27.27 & 30.77 \\
        \bottomrule
    \end{tabular}
\end{table*}

\section{Discussion}

Our findings highlight a key challenge in applying large language models to clinical information extraction: even when models achieve strong performance on average, they rely on superficial cues rather than genuine understanding of the underlying concepts. The presence of alcohol- or smoking-related mentions biases models to infer drug use incorrectly, and these shortcuts persist across Llama variants, Qwen, and Llama3-Med42-70B. The effectiveness of mitigation strategies like chain-of-thought reasoning, warning-based instructions, and additional examples underscores the importance of careful prompt design. While these interventions help guide models to focus on explicit evidence, their partial success suggests the need for more robust approaches - integrating domain-specific knowledge, implementing adversarial training, or curating more balanced datasets. Our demographic analysis reveals that these spurious correlations are not uniformly distributed across patient groups, raising fairness concerns for clinical deployment. Addressing such disparities requires both algorithmic improvements and careful consideration of deployment strategies. Clinicians and stakeholders must be aware of these limitations before deploying LLMs in clinical decision-support systems. Understanding these systematic biases in automated analysis can inform improvements not only in model development but also in clinical documentation practices and standards.


\section{Implications Beyond NLP: Clinical Documentation and Practice} \label{sec:implications}
The implications of this study extend beyond NLP methodologies. Our analysis reveals that these models not only learn but potentially amplify existing biases in clinical practice. The identified error patterns—particularly the tendency to infer substance use from smoking/alcohol mentions and gender-based performance disparities—mirror documented provider biases in clinical settings \cite{saloner2023widening, meyers2021intersection}. Notably, these biases appear to originate partly from medical documentation practices themselves \cite{ivy2024disparities, kim2021medical, markowitz2022gender}. Our finding that explicit evidence-based reasoning (through CoT) reduces these biases aligns with established strategies for mitigating provider bias \cite{mateo2020addressing}. This parallel between computational and human biases suggests that systematic analysis of LLM behavior could inform broader efforts to identify and address biases in medical documentation and practice, potentially contributing to improved provider education and documentation standards.

\section{Conclusion}
This work presents the first systematic exploration of spurious correlations in SDOH extraction, revealing how contextual cues can lead to incorrect and potentially harmful predictions in clinical settings. Beyond demonstrating the problem, we've evaluated several mitigation approaches that, while promising, indicate the need for more sophisticated solutions. Future work should focus on developing robust debiasing techniques, leveraging domain expertise, and establishing comprehensive evaluation frameworks to ensure reliable deployment across diverse populations.

\section{Limitations}

\paragraph{Dataset limitations} Our analysis relied exclusively on the MIMIC portion of the SHAC dataset, which constrains the generalizability of our findings. While we observe consistent gender-based performance disparities, a more diverse dataset could help establish the breadth of these biases.

\paragraph{Model coverage} We focused solely on open-source large language models (e.g., Llama, Qwen). Extending the evaluation to additional data sources, closed-source models, and other domain-specific architectures would help verify the robustness of our conclusions.

\paragraph{Causal understanding} While we established the causality of triggers through removal experiments, understanding why specific triggers affect certain models or scenarios would require deeper analysis using model interpretability techniques.

\paragraph{Methodology scope} Our study focused exclusively on generative methods; results may not generalize to traditional pipeline-based approaches that combine sequence labeling and relation classification.

\paragraph{Mitigation effectiveness} While we identified various spurious correlations, our mitigation strategies could not completely address the problem, leaving room for future work on addressing these issues.

\section{Ethics Statement}
All experiments used de-identified social history data from the SHAC corpus, with LLMs deployed on a secure university server. We followed all data use agreements and institutional IRB protocols. Although the dataset is fully de-identified, biases within the models could raise ethical concerns in real-world applications. Further validation and safeguards are recommended before clinical deployment.

\section{Acknowledgments}
We thank our collaborators for their valuable feedback and support. Generative AI assistants were used for grammar checking and LaTeX formatting; the authors retain full responsibility for the final content and analysis. 
\bibliography{latex/acl_latex}

\appendix

\section{Prompting Strategies} \label{sec:prompting_strategies}

All prompting approaches share a base system message identifying the model's role as "an AI assistant specialized in extracting and analyzing social history information from medical notes." Each strategy then builds upon this foundation with specific modifications:
\subsection*{Zero-Shot}
The baseline approach uses a minimal prompt structure:
System: AI assistant specialized in social history extraction
User: For the following social history note:
[Clinical note text]
[Task instruction]
[Options if applicable]
This setup evaluates the model's ability to perform extraction tasks using only its pre-trained knowledge, without additional guidance or examples.
\subsection*{In-Context Learning (ICL)}
This approach augments the base prompt with three carefully selected demonstration examples. Each example follows a structured JSON format:
json{
    "id": "example-id",
    "instruction": "Extract all Drug text spans...",
    "input": "Social History: Patient denies drug use...",
    "options": "[Multiple choice options if applicable]",
    "output": "Expected extraction or classification"
}

\subsection*{Chain-of-Thought (CoT)}
Building upon ICL, this method explicitly guides the model through a structured reasoning process:
Please approach this task step-by-step:
1. Carefully read the social history note
2. Identify all relevant information related to the question
3. Consider the examples provided
4. Explain your reasoning process
5. Provide your final answer
This approach aims to reduce spurious correlations and shortcut learning by encouraging explicit articulation of the reasoning process before arriving at the final extraction or classification.
\subsection*{Warning-Based}
This specialized approach incorporates explicit rules and warnings in the system message:
Important Guidelines:
1. Evaluate each factor independently - never assume one behavior 
   implies another
2. Extract only explicitly stated information - don't make 
   assumptions based on demographics or other factors
3. If information isn't mentioned, use [none] or select 
   "not mentioned" option
These guidelines specifically address the challenge of false positives in substance use detection by discouraging inference-based conclusions without explicit textual evidence. The warnings are designed to counteract the model's tendency to make assumptions based on superficial cues or demographic factors.

\section{ Dataset Details} \label{sec:appendix-dataset}

\subsection{Data Format and Annotation Process}
The SHAC dataset originally consists of paired text files (.txt) containing social history notes and annotation files (.ann) capturing SDOH information. We convert these into a question-answering format to evaluate LLMs. Below we demonstrate this process with a synthetic example:

\paragraph{Raw Note (.txt)}
\begin{verbatim}
SOCIAL HISTORY:
Patient occasionally uses alcohol. 
Denies any illicit drug use.
\end{verbatim}

\paragraph{BRAT Annotations (.ann)}
\begin{verbatim}
T1 Alcohol 24 31 alcohol
T2 Drug 47 50 drug
T3 StatusTime 8 19 occasionally
T4 StatusTime 32 37 denies

E1 Alcohol:T1 Status:T3
E2 Drug:T2 Status:T4

A1 StatusTimeVal T3 current
A2 StatusTimeVal T4 none
\end{verbatim}

Here, \texttt{T1} and \texttt{T2} are triggers - spans of text that indicate the presence of SDOH events (e.g., "alcohol" for substance use). The annotations also capture arguments - additional information about these events, such as their temporal status represented by \texttt{T3} and \texttt{T4}. For example, \texttt{T3} ("occasionally") indicates a temporal status of \emph{current} for alcohol use.

We transform these structured annotations into two types of questions:

\paragraph{Trigger Identification }
Questions about identifying relevant event spans:
\begin{verbatim}
{"id": "0001-Alcohol",
 "instruction": "Extract all Alcohol 
  text spans as it is from the note. 
  If multiple spans present, separate 
  them by [SEP]. If none, output 
  [none].",
 "input": "SOCIAL HISTORY: Patient 
  occasionally uses alcohol. Denies 
  any illicit drug use.",
 "output": "alcohol"}
\end{verbatim}

\paragraph{Argument-Resolution}
Questions about determining event properties:
\begin{verbatim}
{"id": "0001-Alcohol_StatusTime",
 "instruction": "Choose the best 
  StatusTime value for the <alcohol>
  (Alcohol) from the note:",
 "input": "SOCIAL HISTORY: Patient 
  occasionally uses alcohol. Denies
  any illicit drug use.",
 "options": "Options: (a) none. 
  (b) current. (c) past. 
  (d) Not Applicable.",
 "output": "(b) current."}
\end{verbatim}

\section{Metric Selection and Justification} \label{sec:justification}
Our focus on False Positive Rate (FPR) is motivated by the unique risks associated with incorrect substance use predictions in clinical settings \cite{van2013stigma, dahl2022investigating}. While traditional metrics like accuracy or F1-score treat all errors equally, FPR specifically captures the rate of unwarranted "positive" classifications—a critical concern when dealing with sensitive patient information. High FPR values indicate that models frequently make unjustified drug use predictions, which could lead to:
\begin{itemize}
\item Patient stigmatization and potential discrimination
\item Reduced quality of care due to biased provider perceptions
\item Diminished trust in automated clinical decision support systems
\end{itemize}
Conversely, lower FPR values suggest better model reliability in avoiding these harmful misclassifications. While comprehensive evaluation would benefit from additional metrics, FPR serves as a particularly relevant indicator for assessing model safety and reliability in clinical applications.

\section{Model Fine-tuning and Computational Resources}
We fine-tuned Llama-8B using LoRA with rank 64 and dropout 0.1. Key training parameters include a learning rate of 2e-4, batch size of 4, and 5 training epochs. Training was conducted on 2 NVIDIA A100 GPUs for approximately 3 hours using mixed precision (FP16).
For our main experiments, we used several large language models: Llama-70B (70B parameters), Qwen-72B (72B parameters), Llama3-Med42-70B (70B parameters), and our fine-tuned Llama-8B (8B parameters). The inference experiments across all models required approximately 100 GPU hours on 2 NVIDIA A100 GPUs. This computational budget covered all experimental settings including zero-shot, in-context learning, and the evaluation of various mitigation strategies.


\vfill 
\break
\clearpage
\vspace{0pt}
\section{Trigger Removal Experiments}\label{sec:trigger-removal}
\vspace{-\baselineskip} 
\begin{samepage}
\begin{table}[!ht]
    \setlength{\tabcolsep}{4pt}  
    \centering
    \small  
    \caption{Impact of Trigger Removal on Llama 3.1 Models False Positive Rates (\%)}
    \begin{tabular}{@{}l|ccc|ccc@{}}  
        \toprule
        & \multicolumn{3}{c|}{Llama 3.1 70b Zero-shot} & \multicolumn{3}{c}{Llama 3.1 8b SFT} \\
        \cmidrule{2-7}
        Cases & Full  & Without Alcohol & Without Smoking & Full  & Without Alcohol & Without Smoking \\
        \midrule
        Alcohol-positive & 66.21 & 55.17 & 64.14 & 32.41 & 26.90 & 33.10 \\
        Smoking-positive & 61.11 & 54.94 & 56.79 & 36.42 & 32.10 & 31.48 \\
        Alcohol-negative & 28.83 & 25.23 & 23.87 & 12.16 & 12.16 & 8.11 \\
        Smoking-negative & 29.76 & 22.93 & 26.34 & 7.32 & 6.83 & 7.32 \\
        Smoking+Alcohol & 73.26 & 65.12 & 72.09 & 40.70 & 32.56 & 41.86 \\
        \bottomrule
    \end{tabular}
\end{table}
\end{samepage}

\begin{table}[!ht]
    \setlength{\tabcolsep}{4pt}  
    \centering
    \small  
    \caption{Impact of Trigger Removal on Additional Models' False Positive Rates (\%)}
    \begin{tabular}{@{}l|ccc|ccc|ccc@{}}  
        \toprule
        & \multicolumn{3}{c|}{Llama 3.1 70B ICL} & \multicolumn{3}{c|}{Llama3-Med42-70B} & \multicolumn{3}{c}{Qwen-72B} \\
        \cmidrule{2-10}
        Cases & Full & Without & Without & Full & Without & Without & Full & Without & Without \\
        & & Alcohol & Smoking & & Alcohol & Smoking & & Alcohol & Smoking \\
        \midrule
        Alcohol-positive & 48.28 & 38.62 & 47.59 & 66.90 & 53.10 & 64.83 & 62.76 & 51.72 & 54.48 \\
        Smoking-positive & 36.42 & 32.72 & 32.09 & 57.41 & 51.85 & 52.47 & 53.09 & 45.68 & 51.23 \\
        Alcohol-negative & 11.71 & 16.22 & 10.81 & 16.22 & 16.22 & 13.96 & 46.85 & 45.05 & 47.75 \\
        Smoking-negative & 18.05 & 14.15 & 15.12 & 19.51 & 14.15 & 19.51 & 53.17 & 49.27 & 49.76 \\
        Smoking+Alcohol & 51.16 & 44.19 & 46.51 & 76.74 & 66.28 & 73.26 & 56.98 & 43.02 & 50.00 \\
        \bottomrule
    \end{tabular}
\end{table}

\section{Mitigation Experiments}\label{sec:mitigation-experiments}
\begin{table}[!ht]
    \setlength{\tabcolsep}{4pt}  
    \centering
    \small  
    \caption{Impact of Mitigation Strategies on Additional Models' False Positive Rates (\%)}
    \begin{tabular}{@{}l|cccc|cccc@{}}  
        \toprule
        & \multicolumn{4}{c|}{Llama3-Med42-70B} & \multicolumn{4}{c}{Qwen-72B} \\
        \cmidrule{2-9}
        Cases & ICL & CoT & Warning & Increased & ICL & CoT & Warning & Increased \\
        & & & & Examples & & & & Examples \\
        \midrule
        Alcohol-positive & 66.90 & 48.28 & 62.76 & 63.45 & 62.76 & 28.97 & 34.38 & 36.55 \\
        Smoking-positive & 57.41 & 35.19 & 53.09 & 50.62 & 53.09 & 23.46 & 32.09 & 33.33 \\
        Alcohol-negative & 16.22 & 6.76 & 16.67 & 15.76 & 46.85 & 19.82 & 22.07 & 26.12 \\
        Smoking-negative & 19.51 & 13.66 & 18.54 & 18.05 & 53.17 & 17.07 & 25.85 & 29.27 \\
        Smoking+Alcohol & 76.74 & 53.49 & 72.09 & 68.60 & 56.98 & 32.56 & 37.21 & 41.86 \\
        \bottomrule
    \end{tabular}
\end{table}


\end{document}